\documentclass[sigconf]{acmart}

\usepackage{booktabs}
\usepackage{graphicx}
\usepackage{hyphenat}
\graphicspath{{figures/}}

\AtBeginDocument{%
  }

\setcopyright{none}
\renewcommand\footnotetextcopyrightpermission[1]{}
\begin{document}

\title[Post-Training Quantization for Financial Time-Series Forecasting]{Calibration Bets on the Past:\\
Post-Training Quantization for Financial Time-Series Forecasting}


\author{Junyi Ye}
\orcid{0000-0002-6348-5207}
\affiliation{%
  \institution{School of Computing, Montclair State University}
  \city{Montclair}
  \state{New Jersey}
  \country{USA}
}
\email{yej@montclair.edu}

\author{Ivy Gateri Wanjiku}
\affiliation{%
  \institution{School of Computing, Montclair State University}
  \city{Montclair}
  \state{New Jersey}
  \country{USA}
}
\email{wanjikui1@montclair.edu}
\begin{abstract}
Financial forecasting models are typically developed in full precision,
yet production deployment often requires low-precision inference to
reduce memory and computational cost. Post-training quantization (PTQ)
enables such deployment without retraining. However, reliable activation
quantization requires calibration: activation ranges are estimated from
historical data before deployment and then remain fixed during future
inference. The importance of this deployment choice for financial
forecasting remains poorly understood.
We present a systematic study of activation calibration for PTQ in
cross-sectional volatility forecasting on the S\&P 500. Our evaluation
covers seven representative neural architectures, eight walk-forward test years
(2018--2025), and 560 trained models.
We find that activation calibration has little effect at 8 bits but
becomes the primary determinant of predictive performance at 4 bits.
Under default absolute-maximum (abs-max) calibration, static 4-bit
quantization of both weights and activations removes 11--62\% of the
full-precision mean information coefficient in affected architectures.
Replacing abs-max with percentile calibration recovers 53--94\% of this
degradation in the four most affected architectures. 
The preferred activation range also varies across market periods. Narrow ranges
improve resolution under typical market conditions but lose part of
their advantage when test-period market dispersion exceeds the
calibration history. These findings show that activation calibration is
a first-class deployment decision for reliable 4-bit PTQ in financial
forecasting. When substantial degradation remains, 8-bit activations or
weight-only 4-bit quantization provide more robust deployment choices.

\end{abstract}






\pagestyle{plain}

\maketitle

\section{Introduction}

Deploying financial forecasting models in production often requires
low-precision inference to satisfy practical memory, latency, and
hardware constraints
\citep{duarte2018hls4ml,soni2026viability,tqs2026,milets2025patchtst}.
These constraints become more consequential when models repeatedly
generate forecasts across large asset universes, because inference costs
accumulate across assets and update cycles. Post-training quantization
(PTQ) provides an attractive deployment strategy by converting a trained
model into a low-precision model without retraining. However, the
different components of PTQ do not contribute equally to deployment
accuracy. In practice, weight quantization is often robust, whereas
activation quantization depends critically on how activation ranges are
calibrated before deployment.

Unlike model weights, activation ranges cannot be computed directly from
trained parameters. Instead, they are estimated from a historical
calibration dataset before deployment and then remain fixed throughout
future inference. This makes activation calibration a deployment
decision rather than merely a numerical implementation detail. A wide
range preserves rare activation extremes but reduces numerical
resolution for typical activations, whereas a narrow range improves
resolution while increasing clipping. In financial forecasting, this
trade-off is particularly challenging because calibration always relies
on historical market data, whereas deployment occurs under future market
conditions that may differ substantially from the calibration period.

Existing PTQ research has developed effective calibration methods,
outlier-handling techniques, and low-bit quantization schemes for vision
and language models. However, relatively little is known about the
predictive consequences of activation calibration in financial
forecasting. In particular, it remains unclear how sensitive different
forecasting architectures are to activation calibration, how much of the
resulting degradation can be recovered through improved range selection,
and when calibration itself becomes a limiting factor for low-precision
deployment.

To answer these questions, we systematically evaluate activation calibration under a
walk-forward PTQ protocol for cross-sectional volatility forecasting.
The study spans seven representative neural architectures, eight
walk-forward test years from 2018 to 2025, and 560 independently trained
models on an S\&P 500 equity panel. By comparing each quantized
model against its own full-precision checkpoint while varying only the
activation calibration strategy, we isolate the predictive impact of
activation-range selection.

Our study leads to four main observations. First, 8-bit quantization and
4-bit weight-only quantization introduce little predictive degradation
for most architectures. Second, quantizing activations to 4 bits causes the largest losses, with
affected models losing 11--62\% of their full-precision mean information
coefficient under the default range-selection strategy. Third, much of this loss is recoverable
through improved range selection, although recurrent and multi-scale
architectures retain substantial residual sensitivity. Finally, the
preferred activation range itself changes over time, reinforcing the view that activation calibration should be treated as a deployment decision rather
than a fixed preprocessing step.

The contributions of this paper are as follows.

\begin{itemize}
\item To the best of our knowledge, this is the first systematic study of
activation calibration for post-training quantization in financial
forecasting under a walk-forward evaluation protocol.

\item We reveal that 4-bit activation degradation consists of both range-recoverable loss and architecture-dependent residual loss: improved calibration recovers much of the degradation in several architectures, while recurrent and multi-scale models remain substantially sensitive.

\item We show that activation-range preferences evolve across market
periods and translate these observations into practical deployment
guidelines for selecting among percentile W4A4, layer-wise mixed precision,
W8A8, and weight-only W4 quantization.
\end{itemize}

\section{Related Work}
\label{sec:related}

Our work connects two research directions: activation calibration for
PTQ and low-precision inference for
time-series forecasting.

\textbf{Activation calibration for PTQ.}
Weight quantization is generally robust because weight ranges can be
computed directly from trained parameters. Activation quantization is
more challenging because activation ranges must be estimated from a
calibration dataset before deployment. Existing PTQ research has
developed calibration strategies, clipping methods, and quantization-aware
activation functions for low-bit inference
\citep{banner2019post,choi2018pact}. 
More recent work shows that activation outliers dominate many low-bit
failures in large language models, motivating mixed-precision
decomposition \citep{dettmers2022llmint8}, channel-wise scaling
\citep{xiao2023smoothquant}, and activation rotations
\citep{ashkboos2024quarot}.
Calibration data themselves also influence PTQ accuracy even when the
trained model and compression algorithm remain fixed
\citep{williams2024impact}. However, these studies focus primarily on
vision and language models.

\textbf{Low-precision time-series forecasting.}
Low-precision inference has also been studied for sequential and
time-series models through retraining \citep{shin2016fixedpoint} and
quantization-aware training
\citep{fasold2022rnnt,milets2025patchtst}. 
More recently, \citet{tqs2026} show that PTQ sensitivity varies across time-series models and use this variation to motivate mixed-precision allocation. However, these
studies do not systematically investigate activation calibration or
the selection of static activation ranges from calibration data.
Most existing PTQ benchmarks assume calibration and evaluation data
are drawn from the same underlying distribution \citep{yuan2023benchmarking}, whereas financial forecasting requires calibration on historical data followed by deployment on future observations.

To the best of our knowledge, no prior work has systematically
investigated activation calibration for PTQ in financial forecasting.
Our work bridges these two directions by studying how static activation
ranges selected from historical market data affect downstream
forecasting performance under a walk-forward evaluation protocol.

\section{Methodology}
\label{sec:methodology}

\subsection{Low-bit activation quantization}

Post-training quantization (PTQ) converts a trained full-precision
neural network into a low-precision model without retraining. Instead of
representing weights and activations with floating-point numbers, PTQ
approximates them using a small set of discrete values, reducing memory
usage and inference cost.

The reduction in precision becomes particularly significant at low bit
widths. Under the symmetric 4-bit quantizer used in this paper, only 15
representable levels are available to cover the entire numerical range
of an activation tensor. Selecting this range is therefore a central
design decision for reliable low-bit activation quantization.

The range--resolution trade-off can be understood from a standard
symmetric quantizer \citep{jacob2018quantization,nagel2021white}. Let $x$ denote a full-precision scalar and
$\hat{x}$ its quantized--dequantized approximation. For a quantization
range $[-A,A]$, we compute
\begin{equation}
\hat{x}
=
s\cdot
\mathrm{round}
\left(
\frac{\mathrm{clip}(x,-A,A)}{s}
\right),
\qquad
s=\frac{A}{2^{b-1}-1},
\label{eq:qdq}
\end{equation}
where $s$ is the quantization step size for bit width $b$. For the
4-bit case, $s=A/7$.

\begin{figure}[t]
  \centering
  \includegraphics[width=\columnwidth]{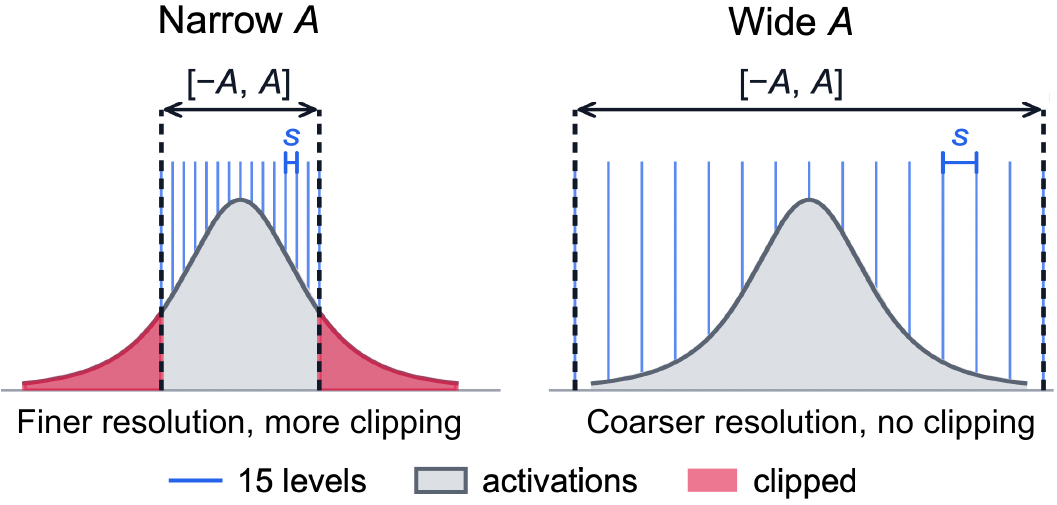}
  \caption{The resolution--range trade-off at 4 bits. Both panels show the
  same activation distribution quantized with the 15 representable
  values; only the range $A$ differs. A narrow $A$ reduces the step size
  $s$ but clips the tails, whereas a wide $A$ eliminates clipping at the
  cost of coarser spacing throughout.}
  \label{fig:schematic}
\end{figure}

Equation~(\ref{eq:qdq}) illustrates the fundamental trade-off. A larger
range reduces clipping by covering more extreme values but increases the
step size, producing larger rounding errors. A smaller range improves
numerical resolution while clipping more activations. The optimal range
therefore balances clipping and rounding errors
\cite{banner2019post}. Even the optimal range generally cannot eliminate
quantization error entirely, leaving a residual error that depends on
both the model and the activation distribution. Figure~\ref{fig:schematic}
illustrates this trade-off by varying only the range parameter $A$ while
holding the activation distribution and quantization precision fixed.

\subsection{Activation calibration}

The quantization range must be determined separately for weights and
activations. Weight tensors are fixed after training, so their ranges
can be computed directly from the trained parameters. Activation values,
however, depend on the input data and therefore require a calibration
dataset before deployment.

We consider static PTQ, a widely used deployment setting in which activation ranges are calibrated once before deployment and then kept fixed throughout subsequent inference. Let $\mathcal C$ denote the calibration set. The activation range is estimated as
\begin{equation}
A
=
\mathrm{stat}
\left(
\{|x|:x\in\mathcal C\}
\right),
\label{eq:static}
\end{equation}
where $\mathrm{stat}(\cdot)$ denotes the calibration statistic. In our
walk-forward evaluation, $\mathcal C$ is sampled from validation year
$Y-1$, and the resulting range is applied unchanged throughout test year
$Y$.

We evaluate two widely used calibration strategies. \textbf{Absolute-maximum (abs-max)
calibration} selects
\[
A=\max\left(\{|x|:x\in\mathcal C\}\right),
\]
thereby covering the full magnitude range observed during calibration. However, a single extreme activation can substantially enlarge the quantization range and reduce numerical resolution for typical activations. \textbf{Percentile calibration} instead selects
\[
A=P_p(\{|x|:x\in\mathcal C\}),
\]
where $P_p$ denotes the $p$-th percentile of activation magnitudes.

Once deployment begins, the activation range remains fixed throughout
the test period. Future activations exceeding the calibrated range are
clipped, even if market conditions differ from those observed during
calibration. The remainder of this paper investigates the predictive
consequences of this fixed calibration strategy for financial
forecasting.

\section{Experimental Setup}\label{sec:setup}

\subsection{Dataset and sample construction}\label{sec:data}

We construct a daily panel of the 501 current S\&P~500 constituents
from June 2008 through December 2025, comprising approximately
2.1 million asset-days. Daily simple returns are computed from Yahoo
Finance\footnote{\url{https://finance.yahoo.com}} adjusted close
prices. Five tickers renamed after the sample period broke their
historical Yahoo Finance identifiers at retrieval time. Their histories
are backfilled from
Tiingo\footnote{\url{https://www.tiingo.com}}, with agreement within
$10^{-3}$ over overlapping periods.


Each stock contributes eleven input features comprising lagged return
statistics, technical indicators, and rolling distribution summaries: daily return: 5-, 10-, and 20-day mean returns; 20-day volatility;
14-day Relative Strength Index (RSI); Moving Average Convergence
Divergence (MACD), computed as the difference of 12- and 26-day
exponential moving averages (EMAs), and its 9-day signal line; and the
20-day maximum, minimum, and skewness of returns.
Each feature is standardized using statistics estimated from the
training split only.

Each input sample consists of a 64-day history of the eleven input
features for a single stock. Models process stocks independently and
produce one volatility forecast for each stock and forecast date.

\subsection{Forecasting task and metric}\label{sec:tasks}
We study cross-sectional prediction of five-day-forward volatility. For
stock $i$ at forecast origin $t$, define
\[
v_{i,t}
=
\mathrm{sd}
\left(
r_{i,t+1},\ldots,r_{i,t+5}
\right).
\]
The supervised target is the within-date cross-sectional z-score of
$\log v_{i,t}$, denoted by $y_{i,t}$.
Cross-sectional volatility forecasting is relevant to risk ranking,
position sizing, and volatility-aware portfolio construction
\citep{moreira2017volatility}. 
Compared with daily return prediction,
volatility provides a stronger predictive signal in this
panel, making it better suited for measuring small
quantization effects.

On each forecast date, the model scores all available stocks. Forecast
skill is measured by the daily information coefficient (IC)
\[
\mathrm{IC}_{t}
=
\mathrm{Spearman}_{i}
\left(
\hat{y}_{i,t},y_{i,t}
\right),
\]
the cross-sectional Spearman correlation between predicted and realized
scores. IC ranges from $-1$ to $1$, with higher values indicating more
accurate cross-sectional ranking and values near zero indicating no
predictive skill. Reported IC values are averaged over all test dates.
Section~\ref{sec:stats} builds on this metric to define quantization
damage as the paired difference between an architecture's
full-precision and quantized IC.

\subsection{Walk-forward protocol}\label{sec:folds}
Evaluation follows a fixed-length walk-forward protocol. For test year
$Y$, training uses the seven years $Y-8$ through $Y-2$, validation uses
year $Y-1$, and testing uses year $Y$. Holding the training-window
length fixed across folds ensures that comparisons across folds are not
confounded by differences in training-window length. The test years
span 2018--2025, producing eight folds. Each of the seven architectures
is trained with ten random seeds per fold, yielding $8 \times 10 \times 7 = 560$ independent training runs.

To prevent information leakage across split boundaries, a sample is
assigned to a split according to the final date of its five-day outcome
window, $t{+}5$, where $t$ is the forecast date, rather than by $t$
itself. This guarantees that every return used to construct a
validation or test target falls strictly within that sample's assigned
split. The 64-day input window, by contrast, is allowed to extend
backward across a split boundary, since those earlier observations are
legitimately available at forecast time. Because input windows only
reach backward while target windows are never split across a boundary,
validation-year early stopping never has access to any test-year
return, whether as a feature or as a label.

Hyperparameters are selected once using a development pseudo-fold with
the same temporal structure (training: 2009--2015; validation: 2016),
and the resulting configuration is frozen for all eight evaluation
folds. 

\subsection{Models and training}\label{sec:models}

We evaluate seven architectures spanning linear, mixer, attention, and
recurrent time-series models:
DLinear \citep{zeng2023dlinear},
TSMixer \citep{chen2023tsmixer},
Time-\\Mixer \citep{wang2024timemixer},
a vanilla Transformer \citep{vaswani2017attention},
PatchTST \citep{nie2023patchtst},
iTransformer \citep{liu2024itransformer}, and
SegRNN \citep{lin2023segrnn}.
TSMixer and TimeMixer use MLP-based mixing. Among the attention-based
models, the vanilla Transformer applies attention directly over raw
time steps without patching or channel tokenization, PatchTST applies
attention over temporal patches, and iTransformer treats input channels
as tokens. SegRNN applies recurrent processing over temporal segments.
For each tunable architecture, we perform an equal-budget grid search
on the development split over
\[
d_{\mathrm{model}}
\in
\{16,32,64,128,256\},
\qquad
\mathrm{dropout}
\in
\{0.1,0.3,0.5\}.
\]
Model depth is fixed at two, with the feed-forward dimension set to
$d_{\mathrm{ff}}=4d_{\mathrm{model}}$ when applicable. Each configuration is trained with three random seeds. The
configuration achieving the lowest average validation mean squared
error (MSE) is selected. The selected
hyperparameters are then fixed for all walk-forward folds. DLinear has no tunable width, depth, or
dropout under its published configuration and is therefore excluded
from the grid search.

All models are trained with AdamW using a batch size of 1024 to minimize the MSE on the standardized target. Training uses early stopping on validation MSE with patience 20 and a maximum of 300 epochs, and the checkpoint with the lowest validation MSE is restored.

\subsection{Classical baselines}\label{sec:baselines}

We include two classical volatility-forecasting baselines as reference
models. The first is a persistence forecast that carries each stock's
trailing five-day realized volatility forward unchanged. The second is
a pooled heterogeneous autoregressive (HAR) model
\citep{corsi2009har} using daily, weekly, and monthly realized
volatility as predictors. HAR is fitted using the training period of
each fold and then applied unchanged to the corresponding validation
and test periods.

\subsection{PTQ protocol}\label{sec:treatments}
\begin{table}[t]
\centering
\caption{Precision settings evaluated in this work. Dynamic INT8 is included only as a reference.}
\label{tab:precision}
\small
\begin{tabular}{lccc}
\toprule
Treatment & Weights & Activations & Activation calibration \\
\midrule
FP32         & FP32 & FP32          & -- \\
W8A8         & INT8 & INT8          & Static \\
W4           & INT4 & FP32          & -- \\
W4A4         & INT4 & INT4          & Static \\
Dynamic INT8 & INT8 & Runtime INT8  & Dynamic \\
\bottomrule
\end{tabular}
\end{table}
Table~\ref{tab:precision} summarizes the evaluated precision settings. We refer to each precision setting as a \emph{treatment}. FP32 serves as the full-precision reference. W4 isolates weight quantization by using INT4 weights with FP32 activations. W8A8 and W4A4 quantize both weights and activations under the static calibration protocol described in Section~\ref{sec:methodology}. Dynamic INT8 is included as a complementary reference. Unlike the static settings, it recomputes activation ranges during inference rather than relying on a fixed calibration range.

The primary experiments focus on W8A8, W4, and W4A4. These settings enable controlled comparisons of weight-only quantization, 8-bit activation quantization, and 4-bit activation quantization under a common static PTQ framework, isolating the quantization configuration and activation-calibration strategy from the underlying execution backend.

\textbf{Activation calibration.}
For W8A8 and W4A4, activation ranges are estimated from 1,024 sequence windows sampled uniformly from the validation split preceding each test year. Unless otherwise specified, the resulting ranges remain fixed throughout the corresponding test year.

\textbf{Quantization coverage.}
We quantize every learned matrix multiplication together with the activation presented to it, including linear projections, convolutional token embeddings, and recurrent matrix products. For SegRNN, the GRU is unrolled and both the step input and hidden-state operands are quantized at every time step. Biases, normalization layers, and elementwise nonlinearities remain in FP32.

For attention-based models, learned projection matrices and their input activations are quantized, whereas activation-only attention products remain in FP32. By quantizing every learned matrix multiplication regardless of architecture, we ensure that the reported fraction of trainable weights covered is comparable across models. Restricting quantization to \texttt{nn.Linear} modules would, for example, leave most of SegRNN's recurrent core unquantized.

\textbf{Execution.}
W8A8, W4A4, and W4 are implemented using simulated quantization. Quantized operands pass through the quantize--dequantize transformation in Eq.~(\ref{eq:qdq}), while matrix multiplication and accumulation are performed in FP32.

\begin{table*}[t]
\centering\small
\caption{Quantization damage ($\Delta$; lower is better) across
architectures. Entries report mean $\pm$ s.d.\ across eight test years
and ten seeds per year; parentheses give damage as a share of FP32 IC
where meaningful. The abs-max, p99.9, and p99 columns are static W4A4
settings differing only in the activation range statistic, with the
lowest-damage setting per architecture shown in bold.}
\label{tab:main}
\resizebox{\textwidth}{!}{

\begin{tabular}{llrrrrrrr}
\toprule
 & & & \multicolumn{3}{c}{Reference settings} &
 \multicolumn{3}{c}{Static W4A4 activation range} \\
\cmidrule(lr){4-6}\cmidrule(lr){7-9}
Model & Family & FP32 IC $\uparrow$ &
$\Delta$Dyn8 $\downarrow$ &
$\Delta$W8A8 $\downarrow$ &
$\Delta$W4 $\downarrow$ &
$\Delta$abs-max $\downarrow$ &
$\Delta$p99.9 $\downarrow$ &
$\Delta$p99 $\downarrow$ \\
\midrule
Persistence & na\"ive & 0.315 & -- & -- & -- & -- & -- & -- \\
HAR & econometric & 0.408 & -- & -- & -- & -- & -- & -- \\
\midrule
DLinear & linear & 0.026$\pm$0.036 & +0.0006 & +0.0007 & +0.0004 & \textbf{+0.006$\pm$0.030} & +0.008$\pm$0.012 & +0.024$\pm$0.018 \\
PatchTST & attention & 0.181$\pm$0.051 & +0.0001 & +0.0000 & -0.0018 & +0.019$\pm$0.041 (11\%) & \textbf{+0.006$\pm$0.028 (3\%)} & +0.008$\pm$0.022 (5\%) \\
iTransformer & attention & 0.197$\pm$0.060 & +0.0004 & +0.0002 & +0.0030 & +0.023$\pm$0.025 (12\%) & +0.016$\pm$0.017 (8\%) & \textbf{+0.015$\pm$0.017 (8\%)} \\
TSMixer & mixer & 0.490$\pm$0.043 & +0.0002 & +0.0002 & +0.0020 & +0.096$\pm$0.041 (20\%) & \textbf{+0.020$\pm$0.012 (4\%)} & +0.027$\pm$0.030 (5\%) \\
Transformer & attention & 0.490$\pm$0.048 & +0.0001 & +0.0003 & +0.0004 & +0.123$\pm$0.066 (25\%) & +0.010$\pm$0.007 (2\%) & \textbf{+0.007$\pm$0.008 (1\%)} \\
SegRNN & recurrent & 0.458$\pm$0.039 & +0.0012 & +0.0006 & +0.0116 & +0.271$\pm$0.068 (59\%) & \textbf{+0.072$\pm$0.030 (16\%)} & +0.107$\pm$0.046 (23\%) \\
TimeMixer & mixer & 0.303$\pm$0.049 & +0.0036 & +0.0084 & +0.0075 & +0.188$\pm$0.134 (62\%) & +0.180$\pm$0.155 (59\%) & \textbf{+0.095$\pm$0.096 (31\%)} \\
\bottomrule
\end{tabular}
}
\end{table*}






\subsection{Quantization damage estimates}\label{sec:stats}
Quantization is applied to a fixed checkpoint, so FP32 and quantized
forecasts are paired on the same stocks and dates. For architecture
$m$, fold $Y$, seed $s$, treatment $q$, and test date $t$, define
\[
\Delta_{m,Y,s,q,t}
=
\mathrm{IC}^{\mathrm{FP32}}_{m,Y,s,t}
-
\mathrm{IC}^{q}_{m,Y,s,t}.
\]
We refer to $\Delta_{m,Y,s,q,t}$ as the \emph{quantization damage}:
positive values indicate that quantization reduces predictive skill
relative to the full-precision checkpoint.

We summarize $\Delta_{m,Y,s,q,t}$ descriptively: it is first averaged
over test dates within each fold-by-seed cell, and we then report the
mean and standard deviation of the resulting fold-by-seed cell means
(Table~\ref{tab:main}, discussed in Section~\ref{sec:results}). These
summaries are purely descriptive and carry no significance markers.
Throughout the paper, we interpret quantization damage through effect
size, comparing it against the HAR and persistence baselines, rather
than through formal hypothesis testing.




\section{Main Results}\label{sec:results}
Table~\ref{tab:main} summarizes the predictive effects of each
quantization setting across 560 walk-forward training runs. The first
three quantized settings (Dynamic INT8, W8A8, and W4) serve as
reference configurations, whereas the remaining columns compare static
W4A4 under different activation-calibration strategies. HAR and
persistence provide classical forecasting baselines that help interpret
the practical magnitude of the observed IC losses.

\textbf{Result 1: Dynamic INT8, W8A8, and W4 preserve nearly all
predictive skill.}
The three reference quantization settings introduce only minor
predictive damage. Under static W8A8, six of seven architectures lose
no more than $0.0007$ mean daily IC. For five of these six, this
corresponds to less than $0.2\%$ of their FP32 signal. The exception is
DLinear, whose near-zero FP32 IC ($0.026$) turns the same small
absolute change into a disproportionately large percentage ($2.7\%$).
Dynamic INT8, which recomputes activation ranges during inference, is
similarly stable: for six architectures, its damage differs from
static W8A8 by less than $0.001$ IC. TimeMixer is the only model with
a clearly visible 8-bit effect, and its loss is smaller under dynamic
than static quantization.

Weight-only W4 is likewise robust in absolute terms. Transformer and
TSMixer lose at most $0.5\%$ of their FP32 IC. DLinear and
iTransformer lose approximately $1.5\%$. As with W8A8, this figure is
inflated for DLinear by its small denominator. No architecture loses
more than $2.5\%$. TimeMixer and SegRNN show the largest weight-only
effects, both around $2.5\%$. These results indicate that neither
8-bit activation quantization nor 4-bit weight quantization accounts
for the substantial degradations observed later.

\textbf{Result 2: Static 4-bit activation quantization causes
substantial predictive losses.}
Under default abs-max calibration, PatchTST loses $11\%$ of its FP32
signal and iTransformer loses $12\%$. The corresponding losses
increase to $20\%$ for TSMixer, $25\%$ for Transformer, and
approximately $60\%$ for both TimeMixer and SegRNN. These differences
do not follow a simple architecture taxonomy: attention-based models
occupy several positions in the damage ordering, as do mixer-based
models, so architecture family alone is insufficient to predict
robustness under static W4A4. 

The practical impact becomes clear when
compared with the classical forecasting baselines. Transformer falls
from an FP32 IC of $0.490$ to $0.367$ under default W4A4, placing it
below HAR. SegRNN falls from $0.458$ to $0.187$, below the persistence
benchmark. The best FP32 neural models exceed HAR by about $0.08$ IC,
while default W4A4 damage is $0.096$ for TSMixer and $0.123$ for
Transformer. For these two architectures, quantization removes an
amount of ranking accuracy comparable to, or larger than, the neural
model's entire advantage over HAR. In contrast, DLinear, whose FP32 IC
is close to zero, changes little under W4A4 and serves as a low-signal
control. 

However, for the architectures that are affected, percentile
calibration (Table~\ref{tab:main}, p99.9 and p99 columns) substantially
reduces damage in some cases but leaves large losses in others. This
pattern motivates Section~\ref{sec:mechanism}, which examines how
activation-range selection shapes W4A4 damage. The magnitude of W4A4
damage also varies substantially across deployment periods, a pattern
that Section~\ref{sec:covid} examines in detail.

\section{Why Default 4-Bit Calibration Fails}\label{sec:mechanism}

\subsection{The range--resolution trade-off}\label{sec:tradeoff}
\begin{figure}[t]
  \centering
  \includegraphics[width=\columnwidth]{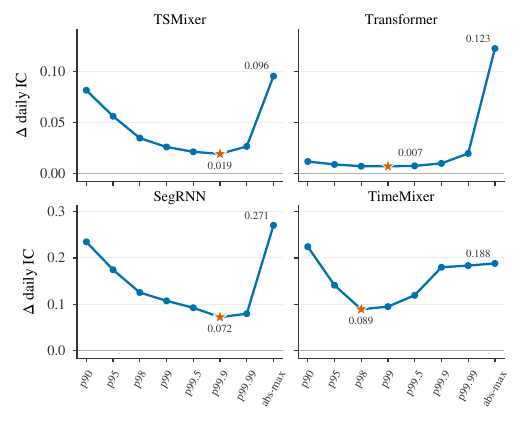}
  \caption{Activation-range sweep for W4A4, averaged over eight
  walk-forward folds and ten seeds per fold. Moving left narrows
  the range, improving in-range resolution while increasing clipping.
  The star marker indicates the lowest-damage setting for each architecture.}
  \label{fig:ucurve}
\end{figure}
A finite-range uniform quantizer creates two errors. Values inside the
range are rounded. Values outside the range are clipped
\citep{banner2019post}. At 4 bits, the activation range decides which
error dominates: a wider range reduces clipping but coarsens all
in-range values, while a narrower range improves in-range resolution
but clips more tail values.

Figure~\ref{fig:ucurve} shows this trade-off. We focus this analysis
on the four architectures with the largest default W4A4 damage:
TSMixer, Transformer, SegRNN, and TimeMixer. For all four, default abs-max calibration
lies on the high-damage side of the curve. Damage first decreases as
the range narrows, reaching a minimum somewhere between p98 and p99.9
depending on the architecture. For TSMixer, Transformer, and SegRNN,
damage then rises sharply over the final step to abs-max. TimeMixer
instead rises more gradually from around p99 onward, without a
comparable late jump.

The curve also separates two cases. For TSMixer and Transformer, most
of the default W4A4 failure is range-recoverable. For SegRNN and
TimeMixer, changing the range helps but does not solve the problem.
Those models remain sensitive even under the best tested range.

\subsection{Recoverable and residual damage}\label{sec:rescue}
Percentile calibration changes only the statistic used to set the
activation range. The trained checkpoint, weights, calibration sample,
and test observations stay fixed. Any improvement can therefore be
attributed to range selection under the same quantizer. Transformer
damage falls from $+0.123$ under abs-max to $+0.007$ at the best tested
percentile, while TSMixer damage falls from $+0.096$ to $+0.019$.

We measure how much of the default abs-max damage can be recovered
simply by switching the calibration statistic. Let
$\Delta_{\text{abs-max}}$ denote an architecture's W4A4 damage under
abs-max calibration, and let $\Delta_p$ denote its damage under
percentile calibration at setting $p$ (e.g., p99, p99.9). The best percentile setting is the one that minimizes $\Delta_p$
across the values tested in Figure~\ref{fig:ucurve}, denoted
$\min_p \Delta_p$.
We define two quantities from this comparison: the
\emph{range-recoverable share} $R$, the fraction of abs-max damage
eliminated by switching to the best percentile, and the
\emph{residual damage} $L$, the damage that remains even under that
best setting. Formally,
\[
R =
\frac{\Delta_{\text{abs-max}}-\min_p\Delta_p}{\Delta_{\text{abs-max}}},
\qquad
L = \min_p\Delta_p.
\]
Because the best percentile here is selected using the same test-year
outcomes it is evaluated on, $R$ and $L$ describe an upper bound on
what range selection could achieve: a real deployment would have to
choose the percentile in advance, using only data available before the
test year begins. Section~\ref{sec:practice} turns these patterns into deployment
guidance and tests whether that guidance survives without foresight of
the test period.

The range-recoverable share is $94\%$ for Transformer, $80\%$ for
TSMixer, $73\%$ for SegRNN, and $53\%$ for TimeMixer. These figures
confirm the pattern in Figure~\ref{fig:ucurve}: Transformer and
TSMixer recover most of their default damage through range selection
alone, while SegRNN and TimeMixer recover less than three-quarters and
just over half, respectively.

\subsection{Layerwise attribution}\label{sec:layerwise}
Range selection is not the only lever for reducing W4A4 damage. A
layerwise check shows that Transformer and TSMixer respond very
differently to precision reallocation at the layer level. For
Transformer, one layer dominates: quantizing only the convolutional
token embedding reproduces $72\%$ of full-model W4A4 damage. Keeping
that layer at 8 bits while quantizing the remaining eligible
operations to 4 bits reduces damage from $+0.123$ to $+0.015$,
essentially eliminating the W4A4 damage. TSMixer has no comparable
single-layer bottleneck: protecting any individual layer leaves
damage near $+0.086$, still large enough to erase most of the neural
model's advantage over HAR.


\section{Market Regime Sensitivity}\label{sec:covid}
Percentile calibration reduces average damage by exchanging some tail
coverage for finer bulk resolution. This section examines how that
trade-off changes when the market a model is tested on differs from
the market it was calibrated on, using the COVID-19 shock of 2020, the
calmer recovery in 2021, and the 2022 inflation and
monetary-tightening episode as cases where this mismatch runs in
opposite directions.

\subsection{Calibration--test mismatch}\label{sec:regimes}
Calibration always uses the most recent available data, so it can lag
behind the market it is meant to prepare for. We show that the
direction of that lag matters: a test year can be either more volatile
or calmer than the year it was calibrated on, and these two directions
damage the model in different ways.

\begin{figure}[t]
  \centering
  \includegraphics[width=\columnwidth]{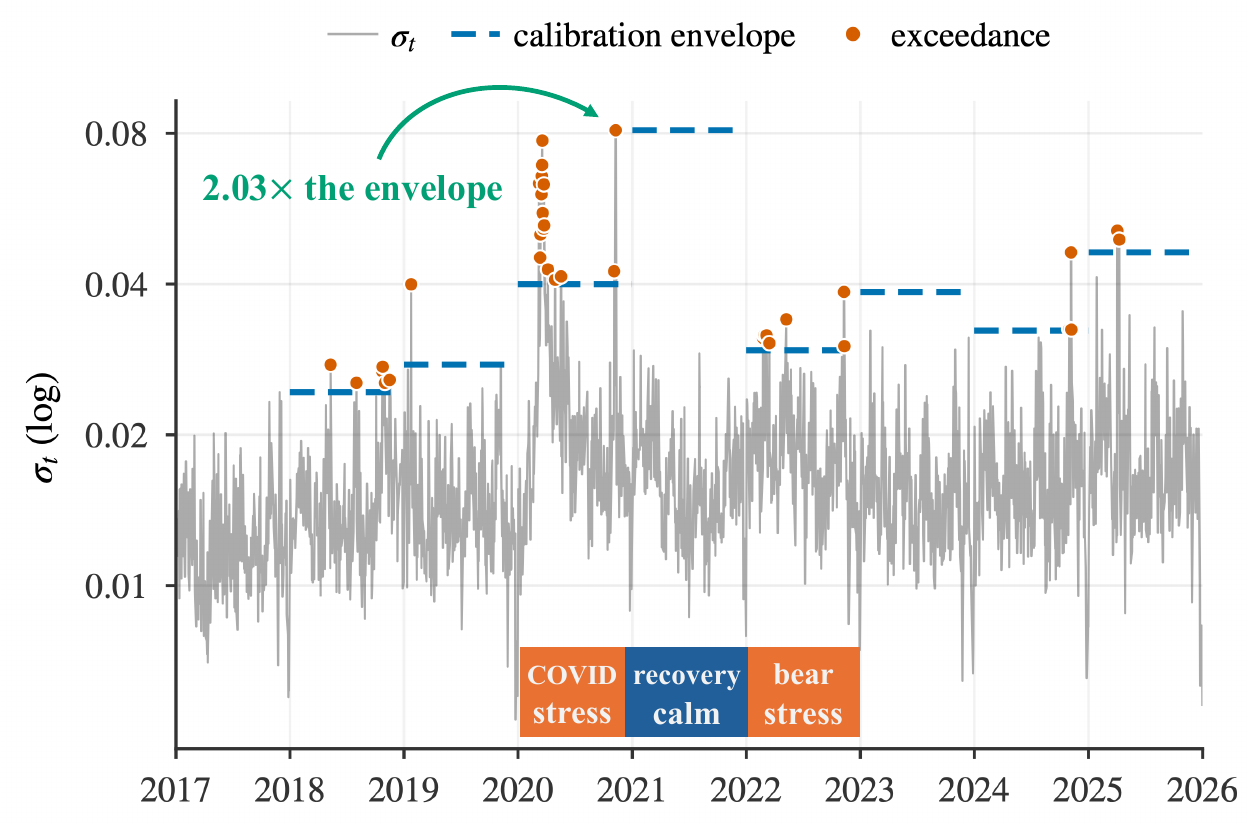}
  \caption{Daily cross-sectional return dispersion $\sigma_t$ (gray)
  against the calibration envelope (dashed blue) for each
  walk-forward fold. Each envelope segment is the maximum $\sigma_t$
  observed during the validation year that precedes it and is held
  fixed while evaluating the following test year (orange markers
  indicate days that exceed it).}
  \label{fig:regime}
\end{figure}

We measure this lag using $\sigma_t$, the daily cross-sectional
standard deviation of stock returns: since the models rank stocks by
future volatility within each date, cross-sectional dispersion is a
direct measure of how unusual that day's ranking environment is. For
each fold, the \emph{calibration envelope} is the largest $\sigma_t$
observed during the preceding validation year. A test day lies outside
the envelope when its dispersion exceeds that maximum.

Figure~\ref{fig:regime} shows that the envelope tracks recent history
rather than a fixed baseline, and that 2020 and 2021 sit on opposite
sides of it. In 2020, the COVID-19 shock pushes dispersion repeatedly
beyond its own envelope: the market moves well beyond what the
preceding year prepared the model for. In 2021, dispersion instead
sits well below the envelope: the market recovers from the 2020 shock,
and cross-sectional dispersion falls back toward typical levels, but
the envelope still reflects the extremes of the year it was calibrated
on. This asymmetry, being calibrated too narrow in one direction and
too wide in the other, is what produces the two failure modes examined
next.

\begin{figure}[t]
  \centering
  \includegraphics[width=\columnwidth]{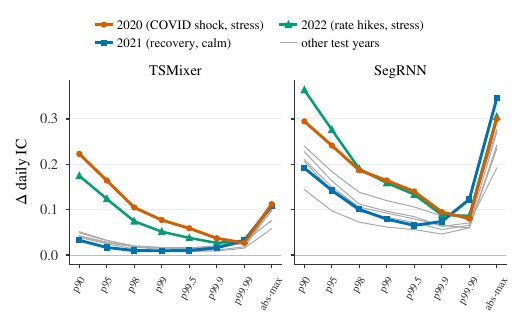}
  \caption{The activation-range sweep of Figure~\ref{fig:ucurve},
  separated by test year for TSMixer and SegRNN, against a baseline of
  all other test years (gray).}
  \label{fig:ucurve_years}
\end{figure}

This mismatch produces two opposite failure modes
(Figure~\ref{fig:ucurve_years}). When the test year is more volatile
than its calibration year, as in 2020 and 2022, narrow percentiles
cost more than usual: they clip tail activations the calibration
period underestimated. When the test year is calmer than its
calibration year, as in 2021, the opposite happens: abs-max becomes
the costliest setting, since it preserves resolution for extremes that
no longer occur. For SegRNN, this reversal is severe enough that the
2021 fold has the largest abs-max damage of any fold in the panel.

This mismatch also inflates the damage that remains even after
choosing the best percentile. Stress years produce more extreme tail
activations, and no single percentile can fully cover them without
sacrificing resolution elsewhere, so clipping-driven damage persists
even under the best tested range. For TSMixer, residual damage in 2020 and 2022 is roughly twice that of
2021 and the remaining calmer folds.

\subsection{Matched recalibration}\label{sec:oracle}

We ask what would have happened to the 2020 fold if its calibration
period had not been mismatched with its test period. The 2020 fold
differs from other folds in several respects simultaneously: training
sample, checkpoint, calibration period, and test period, so it is not
possible to attribute its excess damage to calibration--test mismatch
alone from the original comparison. To isolate this factor, we
construct a matched comparison in which the 2020 fold is recalibrated
on 2020 data itself, rather than on 2019, while the checkpoint, test
days, range statistic, and sampling procedure remain unchanged.

This comparison is diagnostic rather than deployable, since it uses
2020 data that would not have been available before the year began. It
answers a narrower question: how much of the 2020 excess damage under
p99 calibration would have disappeared had calibration matched the
test period, holding all other factors fixed.

Matched p99 recalibration reduces TSMixer damage from $+0.077$ to
$+0.043$, a reduction of $44\%$, and SegRNN damage from $+0.164$ to
$+0.119$, a reduction of $27\%$. Calibration--test mismatch therefore
accounts for roughly a quarter to half of the 2020 excess damage under
p99, depending on the architecture. The remaining loss persists even
when calibration matches the test year exactly, indicating that it
comes from clipping itself: a fixed percentile still cannot cover the
extreme tail activations that a stress year like 2020 produces, no
matter which year's data set that percentile uses.

\subsection{The cost of narrow ranges on extreme days}\label{sec:envelope}





\begin{table}[t]
\centering
\small
\setlength{\tabcolsep}{5pt}
\caption{Averaged daily IC damage relative to FP32 inside and
outside the calibration envelope. Positive values indicate lower IC
after quantization. The ratio columns report outside/inside damage
under p99 and abs-max calibration.}
\label{tab:dayattrib}
\begin{tabular}{lrrrrrr}
\toprule
 & \multicolumn{3}{c}{$\Delta$p99} & \multicolumn{3}{c}{$\Delta$abs-max} \\
\cmidrule(lr){2-4}\cmidrule(lr){5-7}
Model & Inside & Outside & Ratio & Inside & Outside & Ratio \\
\midrule
TimeMixer   & +0.094 & +0.140 & 1.5 & +0.187 & +0.266 & 1.4 \\
TSMixer     & +0.026 & +0.056 & 2.2 & +0.096 & +0.088 & 0.9 \\
SegRNN      & +0.104 & +0.260 & 2.5 & +0.270 & +0.285 & 1.1 \\
Transformer & +0.007 & +0.025 & 3.6 & +0.123 & +0.095 & 0.8 \\
\bottomrule
\end{tabular}
\end{table}

Percentile calibration lowers average damage, but Table~\ref{tab:dayattrib}
shows that this saving is conditional, not free. Under p99, damage
outside the calibration envelope is $1.5$--$3.6\times$ larger than
inside, for all four architectures. The narrow range that helps on typical days clips more heavily once
conditions turn extreme. Its advantage disappears when
reliability matters most, on the unusual market days a deployed model
is least prepared for. Abs-max does not carry this cost. It is already wide enough to cover
extremes, so its damage outside the envelope is no higher than inside
for TSMixer and Transformer, and modestly higher for TimeMixer and SegRNN.





\section{Deployment Implications and Limitations}\label{sec:practice}

\begin{table}[t]
\centering\small
\caption{Deployment choices implied by the range and residual
diagnostics.}
\label{tab:guidance}
\begin{tabular}{ll}
\toprule
Observed pattern & Suggested setting \\
\midrule
No meaningful W4A4 damage
    & Keep the default range \\
Most damage disappears after range tuning
    & Percentile W4A4 \\
One layer accounts for most damage
    & Keep that layer at 8 bits \\
Large damage remains across tested ranges
    & Dyn8, W8A8, or W4 \\
\midrule
Test conditions exceed calibration history
    & Recalibrate or use 8 bits \\
\bottomrule
\end{tabular}
\end{table}

\textbf{A deployment procedure.} Table~\ref{tab:guidance} summarizes the resulting
deployment procedure. Start from the full-precision checkpoint and
measure default W4A4 damage before deployment. If the damage is
small, no range intervention is needed. If the damage is large, sweep
percentile ranges on data available before deployment and compute
$R$ and $L$; together with the layerwise check, these determine the
setting shown in the table. After deployment, the calibration
envelope provides a monitoring signal: among the architectures most
sensitive to activation range, out-of-envelope days have
$1.5$--$3.6\times$ the typical daily quantization damage under p99
calibration. Such days can trigger recalibration, a more conservative
range such as p99.9, or a fallback to 8-bit activations.

\textbf{Does this guidance survive without hindsight?} The range
sweeps in Section~\ref{sec:mechanism} choose the best percentile using
test-year outcomes, so they are diagnostic rather than deployable. To
test a deployable rule, we repeat the same sweep on the validation
year. For each architecture and fold, we choose the percentile $p$
with the smallest validation-year damage $\Delta_p$, then apply that
percentile to the following test year. We refer to the
retrospectively best test-year percentile as the \emph{oracle}
percentile.

This validation rule preserves the coarse guidance from $R$ and $L$
for all seven architectures. Across the four swept architectures, the validation-selected percentile has mean regret of $0.007$ IC and median regret of $0.002$ IC relative to the oracle, and matches or improves on a fixed p99 rule in 27 of the 32 architecture–fold cells.
The largest failure occurs for SegRNN in the 2021 fold. Its percentile
is selected from the 2020 validation year, a COVID-stress period, and
then applied to the calmer 2021 test year. The resulting regret is
$0.057$ IC.

\textbf{Limitations and future work.}
Reported IC levels are survivors-hip-tilted because the panel uses
current S\&P~500 constituents. This bias is shared across both
conditions, since quantized models are compared with their own
full-precision checkpoints on identical stocks and dates.
Quantization is simulated in FP32, capturing the predictive cost of
range selection but not backend-level efficiency or accumulation
effects. We also study only symmetric round-to-nearest quantization, in its standard hardware-friendly layout: per-tensor activation ranges with per-channel weight scales. Stronger PTQ methods~\cite{ashkboos2024quarot,xiao2023smoothquant} may shrink the residual damage of SegRNN and TimeMixer, an open question this framework can answer. Finally, the eight test years
contain only two major stress episodes, and skill is measured by IC
alone. Hardware validation, other markets, and portfolio-level
evaluation remain future work. These limitations bound the
generality of the specific numbers reported here, not the central
finding: activation calibration is not a minor implementation detail
but a first-order determinant of 4-bit deployment performance.
\section{Conclusion}
Post-training quantization is not uniformly risky for financial
time-series forecasting. 8-bit quantization and weight-only 4-bit
quantization generally cause little damage. The large failures come
from static 4-bit activation quantization, where default abs-max
calibration removes a substantial share of predictive signal.

Activation-range selection is a major source of this damage. Percentile
calibration recovers most of the loss, though recurrent and
multi-scale architectures retain substantial residual damage, and the
preferred range shifts across market regimes. Where a single layer
dominates the damage, protecting it at 8 bits recovers what range
selection alone cannot.

Calibration should therefore be treated as part of the deployment
policy, not a fixed preprocessing step: measure default W4A4 damage,
sweep the activation range, and fall back to 8-bit activations,
weight-only W4, or a layerwise exception when residual damage
remains large. This guidance works without foresight of the
deployment period.

\bibliographystyle{ACM-Reference-Format}
\bibliography{references}

@article{corsi2009har,
 title={A simple approximate long-memory model of realized volatility},
  author={Corsi, Fulvio},
  journal={Journal of financial econometrics},
  volume={7},
  number={2},
  pages={174--196},
  year={2009},
  publisher={Oxford University Press}
}

@misc{dettmers2022llmint8,
    title={LLM.int8(): 8-bit Matrix Multiplication for Transformers at Scale}, 
    author={Tim Dettmers and Mike Lewis and Younes Belkada and Luke Zettlemoyer},
    year={2022},
    eprint={2208.07339},
    archivePrefix={arXiv},
    primaryClass={cs.LG},
    url={https://arxiv.org/abs/2208.07339}, 
}

@inproceedings{xiao2023smoothquant,
  title={SmoothQuant: Accurate and efficient post-training quantization for large language models},
  author={Xiao, Guangxuan and Lin, Ji and Seznec, Mickael and Wu, Hao and Demouth, Julien and Han, Song},
  booktitle={International conference on machine learning},
  pages={38087--38099},
  year={2023},
  organization={PMLR}
}

@article{moreira2017volatility,
  title={Volatility-managed portfolios},
  author={Moreira, Alan and Muir, Tyler},
  journal={The Journal of Finance},
  volume={72},
  number={4},
  pages={1611--1644},
  year={2017},
  publisher={Wiley Online Library}
}

@inproceedings{nie2023patchtst,
    title={A time series is worth 64 words: Long-term forecasting with transformers},
  author={Nie, Yuqi and Nguyen, Nam H and Sinthong, Phanwadee and Kalagnanam, Jayant},
  booktitle={ICLR}, year={2023}
}

@inproceedings{liu2024itransformer,
 title={{iTransformer}: Inverted transformers are effective for time series forecasting},
  author={Liu, Yong and Hu, Tengge and Zhang, Haoran and Wu, Haixu and Wang, Shiyu and Ma, Lintao and Long, Mingsheng},
  booktitle={ICLR}, year={2024}
}

@inproceedings{zeng2023dlinear,
  title={Are transformers effective for time series forecasting?},
  author={Zeng, Ailing and Chen, Muxi and Zhang, Lei and Xu, Qiang},
  booktitle={Proceedings of the AAAI conference on artificial intelligence},
  volume={37},
  number={9},
  pages={11121--11128},
  year={2023}
}

@article{tqs2026,
  title={Quantizing time-series models as dynamical systems: Trajectory-based quantization sensitivity score},
  author={Pavlova, Mariya and Zhu, Harrison Bo Hua and Vitanova, Lidia and Semenova, Elizaveta and Li, Yingzhen},
  journal={arXiv preprint arXiv:2606.13300},
  year={2026}
}

@inproceedings{milets2025patchtst,
   title={Scaled FP32 and Quantization-aware Training of PatchTST for Efficient Time Series Forecasting},
  author={Blair, Lorson and Buhler, Jeremy and El Maghraoui, Kaoutar and Carothers, Christopher D and Wang, Naigang and Murray, Jordan},
  booktitle={Proceedings of the 11th Mining and Learning from Time Series Workshop (MILETS 2025)},
  year={2025}
}

@article{choi2018pact,
  title={{PACT}: Parameterized clipping activation for quantized neural networks},
  author={Choi, Jungwook and Wang, Zhuo and Venkataramani, Swagath and Chuang, Pierce I-Jen and Srinivasan, Vijayalakshmi and Gopalakrishnan, Kailash},
  journal={arXiv preprint arXiv:1805.06085},
  year={2018}
}

@inproceedings{shin2016fixedpoint,
  title={Fixed-point performance analysis of recurrent neural networks},
  author={Shin, Sungho and Hwang, Kyuyeon and Sung, Wonyong},
  booktitle={2016 IEEE International Conference on Acoustics, Speech and Signal Processing (ICASSP)},
  pages={976--980},
  year={2016},
  organization={IEEE}
}

@inproceedings{fasold2022rnnt,
 title={Accelerating Inference and Language Model Fusion of Recurrent Neural Network Transducers via End-to-End 4-Bit Quantization},
  author={Fasoli, Andrea and Chen, Chia-Yu and Serrano, Mauricio and Venkataramani, Swagath and Saon, George and Cui, Xiaodong and Kingsbury, Brian and Gopalakrishnan, Kailash},
  booktitle={Proceedings of Interspeech 2022},
  year={2022},
  doi={10.21437/Interspeech.2022-413}
}

@inproceedings{jacob2018quantization,
  title={Quantization and training of neural networks for efficient integer-arithmetic-only inference},
  author={Jacob, Benoit and Kligys, Skirmantas and Chen, Bo and Zhu, Menglong and Tang, Matthew and Howard, Andrew and Adam, Hartwig and Kalenichenko, Dmitry},
  booktitle={Proceedings of the IEEE conference on computer vision and pattern recognition},
  pages={2704--2713},
  year={2018}
}

@article{nagel2021white,
  title   = {A White Paper on Neural Network Quantization},
  author  = {Nagel, Markus and Fournarakis, Marios and Amjad, Rana Ali and Bondarenko, Yelysei and van Baalen, Mart and Blankevoort, Tijmen},
  journal = {arXiv preprint arXiv:2106.08295},
  year    = {2021},
}

@article{ashkboos2024quarot,
  title={QuaRot: Outlier-free 4-bit inference in rotated LLMs},
  author={Ashkboos, Saleh and Mohtashami, Amirkeivan and Croci, Maximilian L and Li, Bo and Cameron, Pashmina and Jaggi, Martin and Alistarh, Dan and Hoefler, Torsten and Hensman, James},
  journal={Advances in Neural Information Processing Systems},
  volume={37},
  pages={100213--100240},
  year={2024}
}

@article{chen2023tsmixer,
  title={{TSMixer}: An All-{MLP} Architecture for Time Series Forecasting},
  author={Chen, Si-An and Li, Chun-Liang and Yoder, Nate and Arik, Sercan O. and Pfister, Tomas},
  journal={Transactions on Machine Learning Research},
  year={2023}
}

@inproceedings{wang2024timemixer,
  title={TimeMixer: Decomposable Multiscale Mixing for Time Series Forecasting},
  author={Shiyu Wang and Haixu Wu and Xiaoming Shi and Tengge Hu and Huakun Luo and Lintao Ma and James Y. Zhang and Zhou Jun},
  booktitle={The Twelfth International Conference on Learning Representations},
  year={2024},
}

@article{lin2023segrnn,
  title={Segrnn: Segment recurrent neural network for long-term time series forecasting},
  author={Lin, Shengsheng and Lin, Weiwei and Wu, Wentai and Zhao, Feiyu and Mo, Ruichao and Zhang, Haotong},
  journal={IEEE Internet of Things Journal},
  year={2026},
  volume={13},
  number={5},
  pages={9861-9871},
  doi={10.1109/JIOT.2025.3647705},
  publisher={IEEE}
}

@article{vaswani2017attention,
  title={Attention is all you need},
  author={Vaswani, Ashish and Shazeer, Noam and Parmar, Niki and Uszkoreit, Jakob and Jones, Llion and Gomez, Aidan N and Kaiser, {\L}ukasz and Polosukhin, Illia},
  journal={Advances in neural information processing systems},
  volume={30},
  year={2017}
}

@inproceedings{williams2024impact,
  title={On the impact of calibration data in post-training quantization and pruning},
  author={Williams, Miles and Aletras, Nikolaos},
  booktitle={Proceedings of the 62nd Annual Meeting of the Association for Computational Linguistics (Volume 1: Long Papers)},
  pages={10100--10118},
  year={2024}
}

@article{duarte2018hls4ml,
  title={Fast inference of deep neural networks in FPGAs for particle physics},
  author={Duarte, Javier and Han, Song and Harris, Philip and Jindariani, Sergo and Kreinar, Edward and Kreis, Benjamin and Ngadiuba, Jennifer and Pierini, Maurizio and Rivera, Ryan and Tran, Nhan and others},
  journal={Journal of instrumentation},
  volume={13},
  number={07},
  pages={P07027--P07027},
  year={2018}
}

@article{soni2026viability,
  title={Assessing the operational viability of foundation models for time series forecasting},
  author={Soni, Kavin and Das, Debanshu and Guduguntla, Vamshi},
  journal={arXiv preprint arXiv:2605.24381},
  year={2026}
}

@article{banner2019post,
  title={Post training 4-bit quantization of convolutional networks for rapid-deployment},
  author={Banner, Ron and Nahshan, Yury and Soudry, Daniel},
  journal={Advances in neural information processing systems},
  volume={32},
  year={2019}
}

@article{yuan2023benchmarking,
  title={Benchmarking the reliability of post-training quantization: a particular focus on worst-case performance},
  author={Yuan, Zhihang and Liu, Jiawei and Wu, Jiaxiang and Yang, Dawei and Wu, Qiang and Sun, Guangyu and Liu, Wenyu and Wang, Xinggang and Wu, Bingzhe},
  journal={ICML 2023 Workshop on New Frontiers in Adversarial Machine Learning},
  year={2023}
}

\end{document}